\documentclass[10pt, conference, compsocconf]{IEEEtran}
\ifCLASSINFOpdf
\else
\fi
\usepackage{graphicx}
\usepackage{tabularx}
\usepackage{array,multirow,graphicx}
\usepackage{amsmath}
\usepackage{makecell}
\graphicspath{ {} }
\usepackage[numbers]{natbib}
\usepackage{hyperref}       
\usepackage{siunitx}
\newcommand*\rot{\rotatebox{90}}
\begin{document}
%
\title{Real-Time End-to-End Action Detection\\
with Two-Stream Networks}

\author{
\IEEEauthorblockN{Alaaeldin El-Nouby\IEEEauthorrefmark{1}\IEEEauthorrefmark{2}, Graham W.~Taylor\IEEEauthorrefmark{1}\IEEEauthorrefmark{2}\IEEEauthorrefmark{3}}

\IEEEauthorblockA{\IEEEauthorrefmark{1}School of Engineering,
University of Guelph\\
}
\IEEEauthorblockA{\IEEEauthorrefmark{2}Vector Institute for Artificial Intelligence
}
\IEEEauthorblockA{\IEEEauthorrefmark{3}Canadian Institute for Advanced Research
}
\{aelnouby,gwtaylor\}@uoguelph.ca
}


%


\maketitle

\begin{abstract}
Two-stream networks have been very successful for solving the problem of action
detection. However, prior work using two-stream networks train both streams
separately, which prevents the network from exploiting regularities between the
two streams. Moreover, unlike the visual stream, the dominant forms of optical
flow computation typically do not maximally exploit GPU parallelism.
We present a real-time end-to-end trainable two-stream
network for action detection. First, we integrate the optical flow
computation in our framework by using \textit{Flownet2}.
Second, we apply early fusion for the two streams and train the whole pipeline jointly
end-to-end. Finally, for better network initialization, we transfer from the
task of action recognition to action detection by pre-training our framework
using the recently released large-scale \textit{Kinetics} dataset. Our experimental results
show that training the pipeline jointly end-to-end with fine-tuning the
optical
flow for the objective of action detection improves detection performance
significantly. Additionally, we observe an improvement when initializing with
parameters pre-trained using \textit{Kinetics}. Last, we show
that by integrating the optical flow computation, our framework is more
efficient, running at real-time speeds (up to \SI{31}{fps}).

\end{abstract}


%
\IEEEpeerreviewmaketitle

\section{Introduction}
Human spatial action localization and classification in videos are
challenging tasks that are key to better video understanding. Action
detection is especially challenging, as it requires localizing the actor in the
scene, as well as classifying the action. This is done for every frame in a
video with little or no context. In contrast, a related task is action
recognition, which uses signals from all video frames to predict the action. Action
detection has important applications, such as surveillance and human-robot
interaction. However, most current approaches are computationally
expensive and are far from real-time performance, which limits their usage in
real life applications.
%
%

Understanding actions in videos has been an active area of research in recent
years. Following the success of deep convolutional neural networks (CNNs) on
the task of image classification, researchers have used CNNs for the tasks of action
recognition and localization. For image classification, appearance is typically the only
cue available, represented by RGB pixel values. Videos provide an extra
signal: motion. Researchers have worked on many different ways to
model motion cues, including 3D CNNs and recurrent neural networks.
One of the most successful approaches are two-stream networks
\cite{DBLP:journals/corr/SimonyanZ14}, which usually consist of a
spatial network that models appearance, whose input is RGB frames, and a
temporal network that models motion. Optical flow is often chosen as input to
this network; however, other inputs can be used, such as dense trajectories. While
adding the temporal stream often improves the model, it adds complexity, as optical
flow is usually computed using a third party algorithm, which works
separately from the RGB stream. This limits the ability for parallelization and
full utilization of compute resources like GPUs, in addition to memory
overhead. Also, using a third party algorithm prevents the model from
being trainable end-to-end such that the visual and motion pathways cannot
learn to co-ordinate. Finally, as shown in \cite{1712.08416}, optical
flow algorithms optimize the end-point-error (EPE), which does not necessarily
align with the objective for action detection.
%
%

One of the challenges of the action detection task is the absence of large-scale
annotated datasets. This problem forces researchers to work with
relatively shallow architectures or use an architecture that is pre-trained on
the image classification task. Only recently have large-scale datasets
for action recognition emerged, such as \textit{Kinetics}
\cite{DBLP:journals/corr/KayCSZHVVGBNSZ17}. Pre-training on a large-scale
dataset for action recognition should transfer well to the task of action
localization.
%
%
%
%

To the best of our knowledge, all past efforts that used two-stream networks
for action detection trained the two streams separately. The predictions from
both streams were then fused using a fusion algorithm. Training the two streams
separately prevents the model from exploiting dependencies between the
appearance and motion cues. As a downside, training the two networks jointly on
the small action localization dataset might lead to overfitting, as the
model will have a very high capacity when compared to amount of labeled data. However,
pre-training on \textit{Kinetics} should solve this overfitting problem.

In this work, we propose an end-to-end trainable framework for real-time spatial
action detection. Following the advances in real-time object detection, we
build our framework with motivation from \textit{YOLOv2}
\cite{DBLP:conf/cvpr/RedmonF17}, the state-of-the-art real-time object
detector. We generalize its architecture to a two-stream network architecture for
action detection. Instead of training each stream separately, we train both
streams jointly by fusing the final activations from each stream and applying a
convolutional layer to produce the final prediction.

We replace the usual third party algorithms used for computing optical flow
\cite{Farneback:2003:TME:1763974.1764031,Zach:2007:DBA:1771530.1771554,5551149}
with a trainable neural network. We use \textit{Flownet2} \cite{IMKDB17} for optical flow
computation and integrate it in our architecture at the beginning of the
temporal stream. Using \textit{Flownet2} has two advantages: first, the
framework becomes end-to-end trainable. While \textit{Flownet2} is trained to
optimize the EPE, the computed optical flow might not be optimal for the
objective of action detection. Fine-tuning \textit{Flownet2} for the task of
action detection should result in better optical flow for our objective
\cite{1712.08416}. Secondly, while other efforts on action detection usually
use implementations of optical flow algorithms that are totally separate
from the model, integrating the optical flow computation in the network
improves the computational speed of the framework, as it makes better use of
parallelization and reduces the data transfer overhead.

Finally, to address the overfitting problem that may be caused by the use of
small-scale datasets or by training the two streams jointly, we pre-train our model
for the task of action recognition on \textit{Kinetics}. The
pre-trained model is then trained on the task of action detection with a weak
learning rate to preserve a relatively generic feature initialization and to
prevent overfitting.

We test our framework using \textit{UCF-101-24}
\cite{DBLP:journals/corr/abs-1212-0402}, a realistic and challenging dataset for
action localization. We use temporally trimmed videos as our framework does
not yet include temporal localization.


\section{Related Work}

In recent years, deep CNNs have been very successful for computer vision tasks.
Specifically, they have shown great improvements for the tasks of image
classification
\cite{DBLP:journals/corr/SzegedyLJSRAEVR14,DBLP:journals/corr/HeZRS15} and
object detection
\cite{DBLP:journals/corr/GirshickDDM13,DBLP:conf/cvpr/RedmonF17} when compared to
traditional hand-crafted methods. Studying actions in videos has been an
active area of research. Videos provide two types of information: appearance,
which is what exists in static images or individual frames of video, and motion.
Researchers have used different approaches for modeling motion, including
two-stream networks and 3D-CNNs.

Two-stream networks \cite{DBLP:journals/corr/SimonyanZ14} have been one of the
most successful approaches for modeling motion for the tasks of action
recognition and detection. In this approach, the network is designed as two
feed-forward pathways:
a spatial stream for modeling appearance and a temporal stream for
modeling motion. While RGB images are a good representation of appearance
information, optical flow is a good representation for motion. The
spatial and temporal streams take RGB frames and optical flow as inputs,
respectively. Many efforts for solving the action detection problem have followed
this approach. \citet{DBLP:journals/corr/GkioxariM14},
motivated by R-CNNs \cite{DBLP:journals/corr/GirshickDDM13}, use selective
search to find region proposals. They use two separate CNNs (appearance and
motion) for feature extraction. These features are fed to a Support Vector Machine (SVM) to predict
action classes. Region proposals are linked using the Viterbi algorithm.
\citet{DBLP:journals/corr/WeinzaepfelHS15} obtain
frame-level region proposals using EdgeBox
\cite{edge-boxes-locating-object-proposals-from-edges}. The frames are then
linked by tracking high-scoring proposals using a tracking-by-detection approach,
which uses two separate CNNs for modeling appearance and motion, and a SVM
classifier similar to \cite{DBLP:journals/corr/GirshickDDM13}. \citet{Peng2016}, motivated by faster-R-CNN \cite{NIPS2015_5638}, use region
proposal networks (RPNs) to find frame-level region proposals. They use a
motion RPN to obtain high quality proposals and show that it is complementary
to an appearance RPN. Multiple frame optical flows are stacked together and demonstrate
improvement in the motion R-CNN. Region proposals are then linked using the Viterbi
algorithm. Both appearance and motion streams are trained separately. \citet{DBLP:journals/corr/SinghSC16} were the first deep learning-based
approach to address real-time performance for action detection. They
proposed using the single shot detector (SSD) \cite{DBLP:journals/corr/LiuAESR15}, which is a
real-time object detector. They also employ a real-time, but less
accurate, optical flow computation \cite{DBLP:journals/corr/KroegerTDG16}.
Combining these two components, they managed to achieve a rate of \SI{28}{fps}.
They propose a novel greedy algorithm for online incremental action linking
across the temporal dimension. While this work is significantly faster than
previous efforts,
they sacrificed accuracy for speed by using a less accurate flow computation.
\citet{DBLP:journals/corr/KalogeitonWFS17} propose
generalizing the anchor box regression method used by faster R-CNN
\cite{NIPS2015_5638} and SSD \cite{DBLP:journals/corr/LiuAESR15} to anchor
cuboids, which consist of a sequence of bounding boxes over time.  They take a
fixed number of frames as input. Then, feature maps from all frames in this sequence
are used to regress and find scores for anchor cuboids. At test time, the anchor
cuboids are linked to create tubelets, which do not have a fixed temporal
extent. While most methods for solving the action detection problem followed the
two-stream approach, \citet{DBLP:journals/corr/HouCS17} use
3D-CNNs. They suggest generalizing R-CNN to videos by designing a tube CNN (T-CNN). Instead of obtaining frame-level action
proposals and using a post-processing algorithm to link actions temporally to
form action tubes, T-CNN learns the action tubes directly from RGB frames.

Optical flow estimation has been dominated by variational approaches that follow
\cite{Horn81determiningoptical}. Though recently, approaches that use deep CNNs for optical flow estimation
\cite{DBLP:journals/corr/RanjanB16,7410673,IMKDB17} have shown promise.
\textit{Flownet} \cite{7410673} is the first end-to-end trainable deep CNN for
optical flow estimation. It is trained using synthetic data to optimize EPE.
The authors provide two architectures to estimate optical flow. The first is a standard CNN that
takes the concatenated channels from two consequent frames and predicts the
flow directly. The second is a two-stream architecture that attempts to find
a good representation for each image before they are combined by a correlation
layer. However, \textit{Flownet} falls behind other top methods due to
inaccuracies with small displacements present in realistic data.
\textit{Flownet2} \cite{IMKDB17} addresses this problem by introducing a
stacked architecture, which includes a subnetwork that is specialized to small
displacements. It achieves more than 50\% improvement in EPE compared to
\textit{Flownet}. Having a trainable network for estimating optical flow can be
very useful, especially when integrated with other tasks. \citet{1712.08416} studied the integration of trainable optical
flow networks \textit{Flownet} \cite{7410673} and \textit{Spynet}
\cite{DBLP:journals/corr/RanjanB16} on the task of action recognition. They
came up with multiple conclusions that suggest that fine-tuning optical
flow networks for the objective of action recognition consistently demonstrated
improvement.
%
%

\section{Methodology}
\begin{figure*}[!h]
\centering
\includegraphics[scale=0.5]{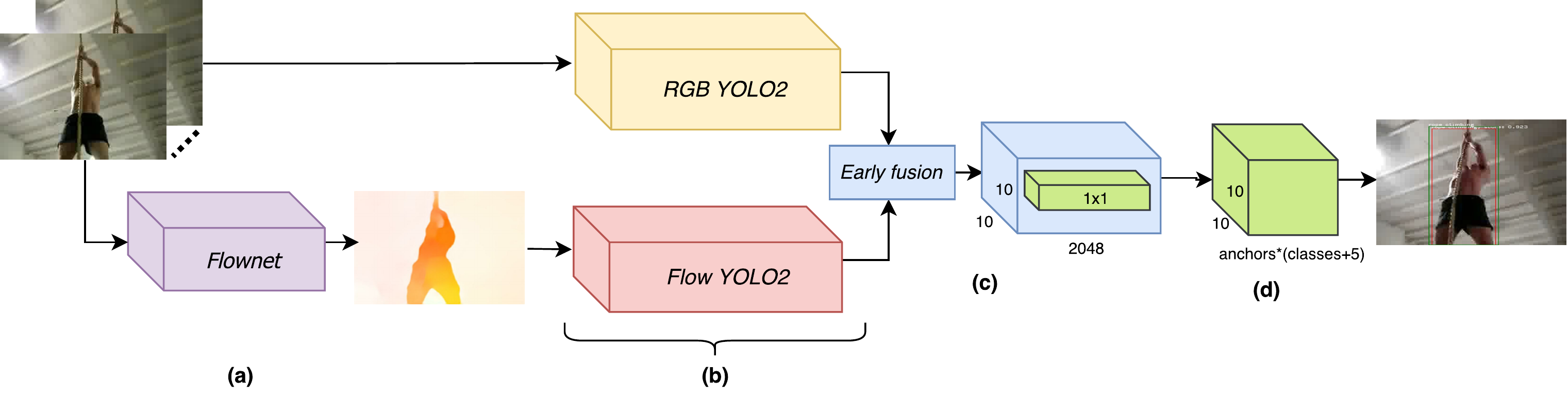}
\caption{Our framework takes a sequence of video frames as input. \textbf{(a)}
\textit{Flownet2} is used to estimate optical flow, which is input to the
 motion stream. \textbf{(b)} The two streams follow the \textit{YOLOv2}
 architecture. \textbf{(c)} We apply early fusion by concatenating the
 activations from both streams channel-wise and then applying a 1x1
 convolutional kernel on the fused activations. \textbf{(d)} Finally, similar
 to \textit{YOLOv2}, the final feature maps are used to regress bounding boxes,
 class scores, and overlap estimates.}
\label{arch}
\end{figure*}

We propose a framework for efficient and accurate action detection, as outlined
in Figure~\ref{arch}. We follow the two-stream network architecture
\cite{DBLP:journals/corr/SimonyanZ14} and integrate optical flow computation in
our framework by using \textit{Flownet2} as input to the motion stream.
We build each stream on \textit{YOLOv2} \cite{DBLP:conf/cvpr/RedmonF17}. In
contrast to previous methods, instead of training each stream separately, we
apply early fusion and train both streams jointly. Finally, the fused feature
maps are used to regress bounding boxes, class scores, and overlap estimates,
similar to \textit{YOLOv2}.

\subsection{Two-Stream YOLOv2 with Early Fusion}

\textit{YOLO} \cite{DBLP:journals/corr/RedmonDGF15} is a real time object
detector. While there have been many successful object detection methods, such as
R-CNN \cite{DBLP:journals/corr/GirshickDDM13}, these methods rely on
extracting region proposals for candidate objects, either by an external
algorithm like Selective Search or EdgeBox, or by a RPN. These proposals are then fed to a CNN to extract features and predict
object classes. In contrast, \textit{YOLO} defines object detection as a
regression problem. A single network predicts both the spatial bounding boxes
and their associated object classes. This design enables end-to-end training and
optimization which allows \textit{YOLO} to run in real-time (\SI{45}{fps}).
Compared to R-CNN \cite{DBLP:journals/corr/GirshickDDM13}, \textit{YOLO} uses
the entire image to predict objects and their locations, meaning that it
encodes appearance as well as contextual information about object classes. This
is very critical for the task of action detection, as context is an extremely
important clue for which action class is present in the scene (e.g., surfing is
associated with sea, skiing is associated with snow). \textit{YOLOv2} is an
improved version of \textit{YOLO}, which adopts the anchor box idea that is
used by R-CNN and SSD. A pass-through layer is added, which brings high
resolution features from early layers on the network to the final low
resolution layers. This layer improves the performance with small-scale objects
that the previous version struggled with. Moreover, \textit{YOLOv2} is even
faster, as it maintains high accuracy with small-scale images. The
fully-connected layer was removed, which makes the network completely
convolutional, reducing the number of parameters. We built our framework on
\textit{YOLOv2}, as it is the best fit for our objective, running at above
real-time speeds while maintaining state-of-the-art accuracy. Moreover, it
encodes better contextual information, which is critical for the task of action
detection. We use the open-source implementation and the pre-trained models
provided by \url{https://github.com/longcw/yolo2-pytorch}.

 In contrast to previous efforts, we train both input streams jointly.
 Training the two streams independently prevents the networks from
 learning complementary features. Associating appearance and motion cues can be
 very useful for identifying the action in
 the scene. We apply early fusion by concatenating the final activations of
 both streams channel-wise. We apply a 1x1 convolutional kernel on top of the
 fused activations. By applying this convolution, we combine the features from
 both streams across each spatial location where there is high correspondence.
 The final activations are used to regress bounding boxes, class scores, and
 overlap estimates, similar to \textit{YOLOv2}.

\subsection{Integrating Flownet}

 Previous two-stream approaches for solving action detection use non-trainable
 optical flow algorithms
 \cite{Farneback:2003:TME:1763974.1764031,Zach:2007:DBA:1771530.1771554,5551149}
  that are completely separate from their detection model. In contrast, we
 integrate optical flow computation in our pipeline. This provides two
 advantages. Firstly, our framework becomes fully trainable end-to-end.
 Fine-tuning optical flow for the task in hand can be very useful. \citet{1712.08416} observe that a CNN trained to optimize the
 EPE might not be the best representative of motion for the task of action
 recognition. They propose fine-tuning the optical flow network for action
 recognition with a weak learning rate and they observe consistent
 improvements. Motivated by this work, we fine-tune \textit{Flownet2} for the
 task of action detection. Secondly, integrating \textit{Flownet2} in our
 pipeline leverages the computational power of GPUs, as all we need is a forward
 pass starting from the video frames to the final detections. Other methods
 usually use publicly available CPU implementations of variational optical
 flow algorithms, which are significantly slower, in addition to data
 transfer overhead. While \cite{DBLP:journals/corr/SinghSC16} uses a less
 accurate, faster optical flow algorithm called DIS-Fast
 \cite{DBLP:journals/corr/KroegerTDG16}, \textit{Flownet2} has architectures
 that are faster with matching quality, or with the same speed with significantly higher
 quality, as shown in Table~\ref{flownet_compare}. We chose to test our model
 with three variations of \textit{Flownet2}. The full-stack architecture
 \textit{Flownet2}, is the most accurate, but slowest architecture.
 \textit{Flownet2-CSS} a less accurate, but faster version. Finally, we test with
 \textit{Flownet2-SD}, a relatively small network that is specialized toward small
 displacements. This model is relatively less accurate than the first two;
 however, it is significantly faster. We use the open-source implementation and
 pre-trained models provided by
 \url{https://github.com/NVIDIA/flownet2-pytorch}.

\newcolumntype{Y}{>{\centering\arraybackslash}X}
\newcolumntype{s}{>{\hsize=.5\hsize}Y}
\def\tabularxcolumn#1{m{#1}}

\begin{table}[!h]
\caption{Average Endpoint Error (AEE) and Runtime comparison of different
variations of \textit{Flownet} and DIS-Fast, as reported in \cite{IMKDB17}}.
\begin{tabularx}{\columnwidth}{|Y|s|s|}
\hline
 Method & \thead{Sintel Final \\  AEE (Train)} & \thead{Runtime \\ (ms per frame)} \\
\hline
DIS-Fast \cite{DBLP:journals/corr/KroegerTDG16} (CPU) & 6.31 & 70 \\
\hline
FlownetS \cite{7410673} (GPU) & 5.45 & \textbf{18} \\
Flownet2-CSS \cite{IMKDB17} (GPU) & 3.55 & 69 \\
Flownet2 \cite{IMKDB17} (GPU) & \textbf{3.14} & 123 \\
\hline
\end{tabularx}

\label{flownet_compare}
\end{table}

\subsection{Pre-Training Using Kinetics}

One of the challenges that researchers face when working on the task of action
detection is the absence of large-scale annotated datasets. Providing
bounding boxes for every frame in every video for a large-scale dataset is an
extremely difficult task. One of the most successful ways to deal with this
kind of problem is through transfer learning. Deep CNN architectures trained on
large-scale image classification datasets like ImageNet \cite{imagenet_cvpr09} have
shown that they can learn features generic enough such that they can be used for
other vision tasks. This suggests that features learned from one task can be
transferred to another. It was also observed that the more similar the two tasks are, the
better the performance after transfer.

After the release of \textit{Kinetics}
\cite{DBLP:journals/corr/KayCSZHVVGBNSZ17}, \citet{DBLP:journals/corr/CarreiraZ17} studied the effect of pre-training
different architectures with \textit{Kinetics} and then used the pre-trained
model to train smaller datasets (e.g., UCF-101, HMDB) for the same task of
action recognition. They report a consistent boost in performance after
pre-training; however the extent of the improvement varies with different
architectures. In this study, the transfer should be optimal, as the target and
source tasks are the same. Previous efforts for solving action detection
usually use network architectures pre-trained on image classification using
ImageNet networks or are pre-trained on the task of object detection using Pascal
VOC \cite{Everingham15}. However, T-CNN \cite{DBLP:journals/corr/HouCS17} uses a
pre-trained C3D model \cite{Tran_2015_ICCV} that is trained using the
\textit{UCF-101} action recognition dataset, which is considerably smaller than
\textit{Kinetics}.

The tasks of action recognition and detection are very similar. In fact, action
recognition can be considered a subtask of action detection. Similarly, action detection and object detection are also related, mainly through the
localization subtask. In order to gain benefit from both tasks and make use of
the large-scale \textit{Kinetics} dataset, we start with \textit{YOLOv2}
architectures for both streams that are pre-trained on object detection using
Pascal VOC. We then train our framework using \textit{Kinetics} with a weak
learning rate in order to preserve some of the features that can help with
localization, while fine-tuning for a different classification task.


\section{Experiments}

We evaluate different variations of our architecture with respect to detection
performance and runtime:

\begin{itemize}
\item \textit{Flownet2} provides improvement in both speed and accuracy.
Therefore, to test the quality of \textit{Flownet2} compared to other accurate
optical flow algorithms, we substitute the method of \citet{5551149}, an accurate but slow optical flow algorithm.

\item Fine-tuning \textit{Flownet2} for the task of action detection produces
optical flow that is a better representation of the action-related motion in
the scene. To validate this idea, we train models with frozen and fine-tuned
\textit{Flownet2} parameters.

\item To investigate transfer learning from the task of activity recognition, we
train models with and without \textit{Kinetics} pre-training. For the models
that were not pre-trained, we use the parameters
trained on object detection using PASCAL VOC.

\item Finally, to have the ability to choose between accuracy and speed, we
substitute \textit{Flownet2} with either
\textit{Flownet2-SD} or \textit{Flownet2-CSS}, observing how they compare
in terms of accuracy and speed to the full-stack estimator.

\end{itemize}

\subsection{Dataset} We use \textit{UCF-101} to test our framework. This is a
dataset that consists of videos for 101 actions in realistic environments
collected from YouTube. This dataset is mainly used for the task of action
recognition. For the action detection task, a subset of 24 actions have been
annotated with bounding boxes, consisting of 3,207 videos.
This is currently the largest dataset available for the task of action
detection. While this dataset includes untrimmed videos, we use the trimmed ones,
as our framework does not include a temporal localization component. We use
split 1 for splitting training and testing data.

\subsection{Evaluation Metric} We use frame mean average precision (f-mAP) to
evaluate our methods. This computes the area under the precision recall curve for
the frame-level detections. A true positive is a detection that has an
intersection over union (IoU) more than a threshold \(\alpha\) with the ground truth,
and the action class is predicted correctly.

\subsection{Implementation Details}

We use PyTorch \cite{paszke2017automatic} for all experimentation. For
\textit{Kinetics} pre-training, we initialize both streams using parameters
trained on PASCAL VOC. We use the SGD
optimizer with a learning rate of 0.0008. We pre-train
\textit{Kinetics} with optical flow from \textit{Flownet2}. We
trained \textit{UCF-101} using the Adam optimizer with a learning rate of
\(5\times10^{-5}\) and batch size of 32. We observed that the Adam optimizer added more stability
when training a multi-task objective. We apply random cropping, HSV
distortion, and horizontal flipping for data augmentation. During training, we
sample two consecutive frames randomly from each sequence. We scale the images
and optical flow to \(320\times320\). For fine-tuning all the \textit{Flownet2}
architectures, we used a learning rate of \(10^{-7}\).
We used the pre-computed Brox \textit{et al.} optical flow provided by
\url{https://github.com/gurkirt/realtime-action-detection}. For testing, we
select the detection box with the highest score in the current frame. We do not
apply any post-processing action linking algorithm.

\section{Results}
\subsection{Ablation Study}

We experiment with different variations of our architecture to show the value
of our proposals. We report the frame mAP at different IoU thresholds for 8
different models in Table.~\ref{abalation}. First, to study the impact of
pre-training using \textit{Kinetics}, we compare it against models
pre-trained using Pascal VOC. We can observe a consistent improvement when
pre-training with \textit{Kinetics}, for both networks trained with Brox,
optical flow, where we notice a 2.5\% gain in frame mAP (0.5 threshold)
or using \textit{Flownet2} where the gain is 4.5\%. The difference
in the gain can be explained by the fact that we pre-trained \textit{Kinetics} using
\textit{Flownet2}. Second, we study the value of fine-tuning \textit{Flownet2}
for the task of action detection. We compare models with frozen and fine-tuned
\textit{Flownet2} parameters. We observe an improvement of 2\% for models
pre-trained with Pascal VOC and 2.5\% for models pre-trained using
\textit{Kinetics}. Combining pre-training with fine-tuning \textit{Flownet2},
we see a gain of \(7\%\). We notice that a model pre-trained with
\textit{Kinetics} and fine-tuned for action detection outperforms all
other variations for all different IoU thresholds.

Finally, we test with \textit{Flownet2-CSS} and \textit{Flownet2-SD} which are
faster, less accurate variations of \textit{Flownet2}. We observe that with
pre-training and fine-tuning, these models outperform the Brox optical flow-trained
model (Brox + VOC), while being significantly faster. We show the AUC curves
for all 8 models we tested in Figure~\ref{auc_curve}.

\newcolumntype{P}[1]{>{\centering\arraybackslash}p{#1}}
{\renewcommand{\arraystretch}{1.2} 
\begin{table}[!h]
\caption{Comparison of variants of our architecture using f-mAP. We test with
	different IoU thresholds \(\alpha\)}.
\begin{tabular}{|m{1.5in}|P{0.4in}|P{0.4in}|P{0.41in}|}
\hline
 Model & \thead{\(\alpha\) = 0.2} & \thead{\(\alpha\) = 0.5} &
 \thead{\(\alpha\) = 0.75} \\
 \hline
Brox + VOC & 77.93 & 70.64 & 32.73 \\
Brox + Kinetics & 80.24 & 73.18 & 33.81 \\
Flownet2 + VOC & 75.43& 66.97 & 28.57 \\
Flownet2 + Kinetics & 79.41 & 71.51 & 32.83 \\
Tuned Flownet2 + VOC & 76.69 & 69.03 & 31.88\\
Tuned Flownet2 + Kinetics & \textbf{81.31} & \textbf{74.07} & \textbf{34.41}\\
Tuned Flownet2-CSS + Kinetics & 79.90 & 72.13 & 32.24 \\
Tuned Flownet2-SD + Kinetics & 78.86 & 71.67 & 33.39 \\
\hline
\end{tabular}
\label{abalation}
\end{table}

\begin{figure}
\includegraphics[trim={1cm, 0cm, 1.5cm, 1.5cm}, 
clip,width=\columnwidth]{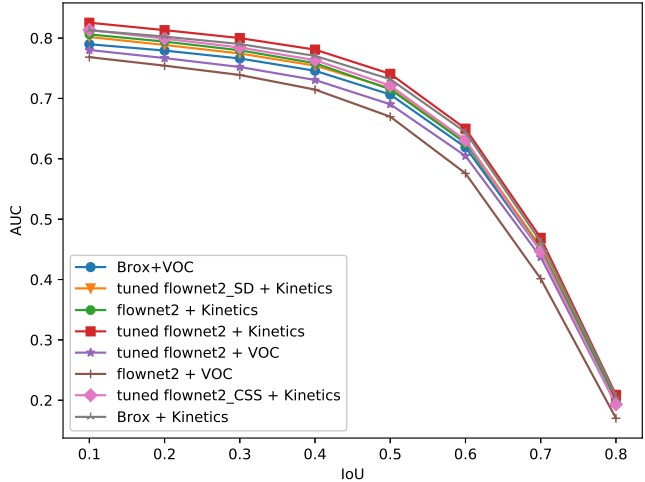}
 \vspace{-2.5em}
\caption{AUC plot for \textit{UCF-101-24} dataset using variations of
our architecture. }
\label{auc_curve}
\end{figure}

\newcolumntype{C}{>{\centering\arraybackslash}m{3.5cm}}
\newcolumntype{k}{>{\hsize=.2\hsize}C}
\newcolumntype{K}{>{\raggedleft\arraybackslash}k}

\begin{figure*}[t]
	\begin{tabularx}{\textwidth}{KCCCC}
		\rot{\scriptsize{Horse Riding}} &  
		\includegraphics[scale=0.15]{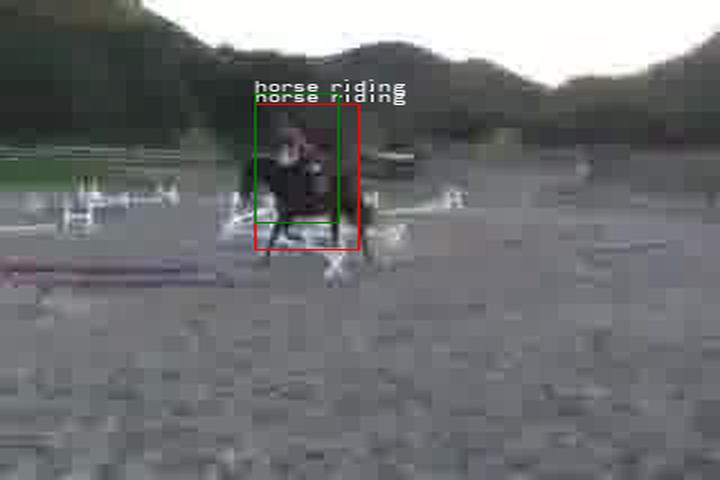}
		&
		\includegraphics[scale=0.15]{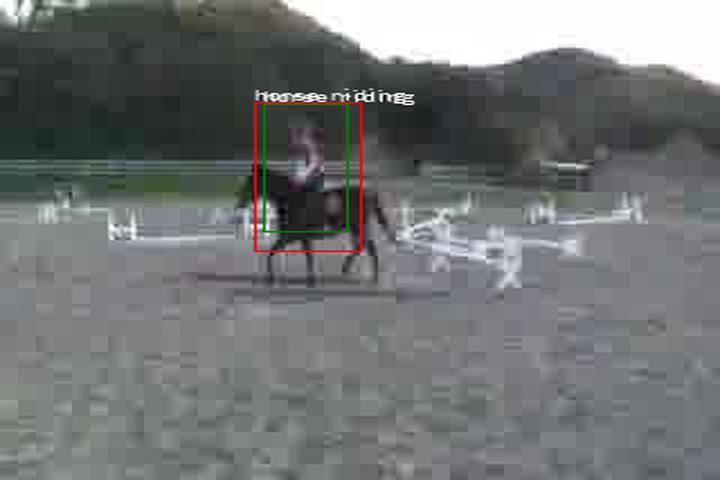} &
		\includegraphics[scale=0.15]{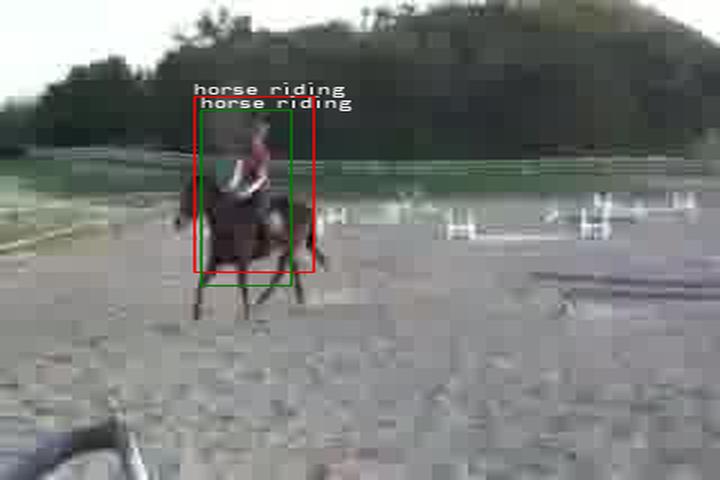} &
		\includegraphics[scale=0.15]{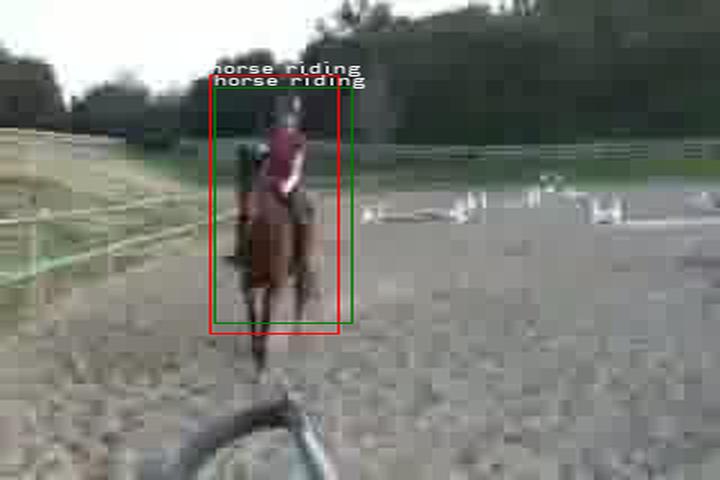} \\
		\rot{\scriptsize{Pole Vaulting}} &  
		\includegraphics[scale=0.15]{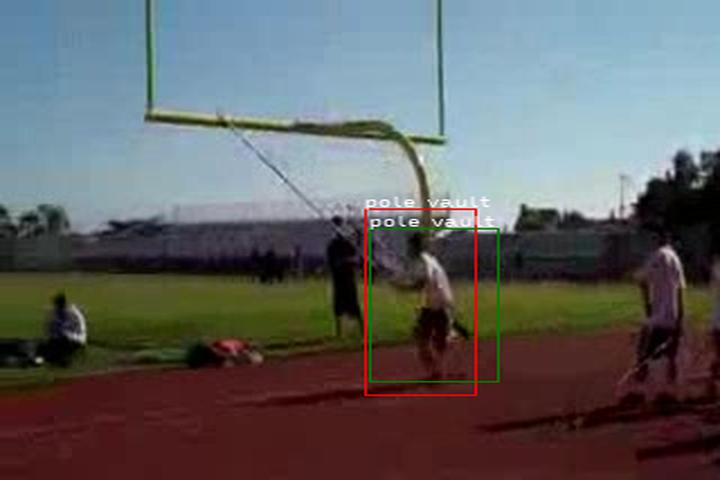}
		&
		\includegraphics[scale=0.15]{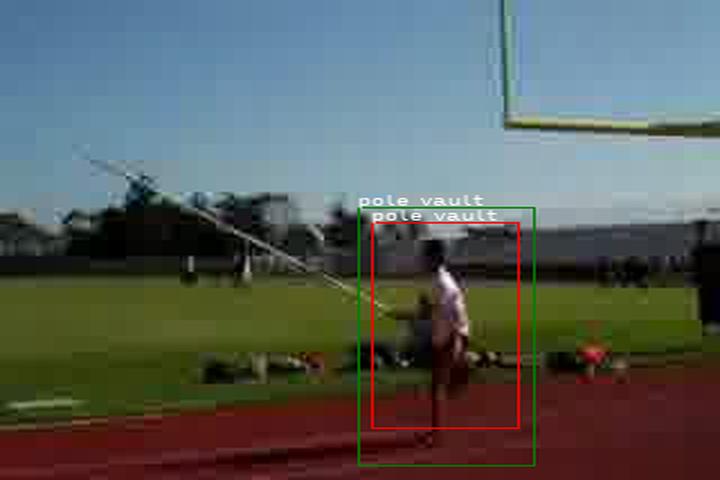} &
		\includegraphics[scale=0.15]{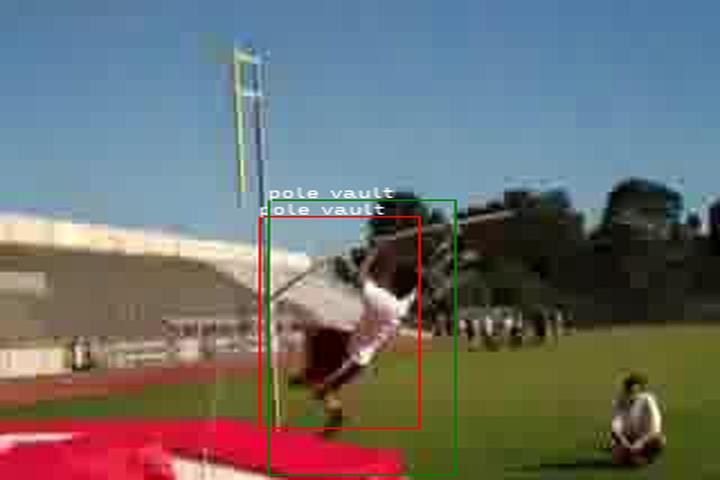} &
		\includegraphics[scale=0.15]{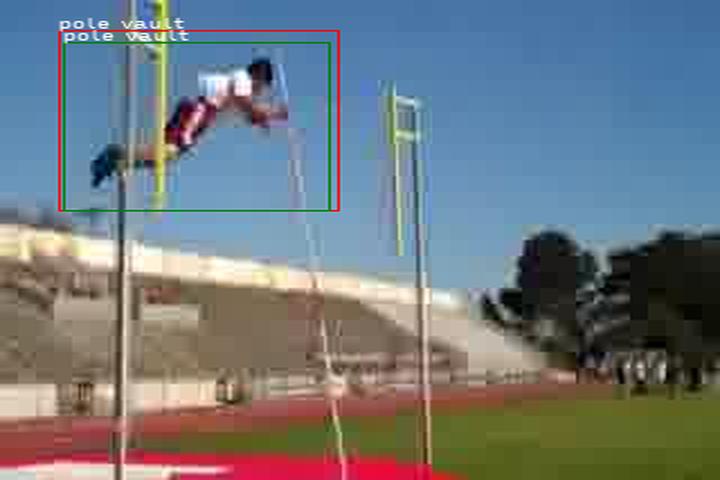} \\
		\rot{\scriptsize{Skiing}} &  \includegraphics[scale=0.15]{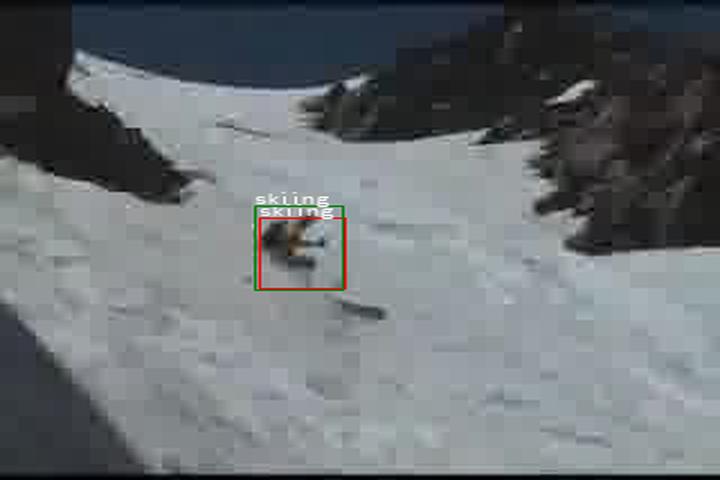} &
		\includegraphics[scale=0.15]{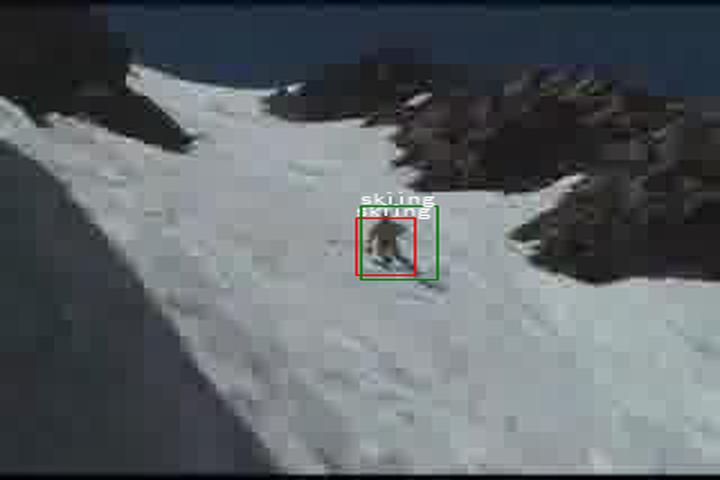} &
		\includegraphics[scale=0.15]{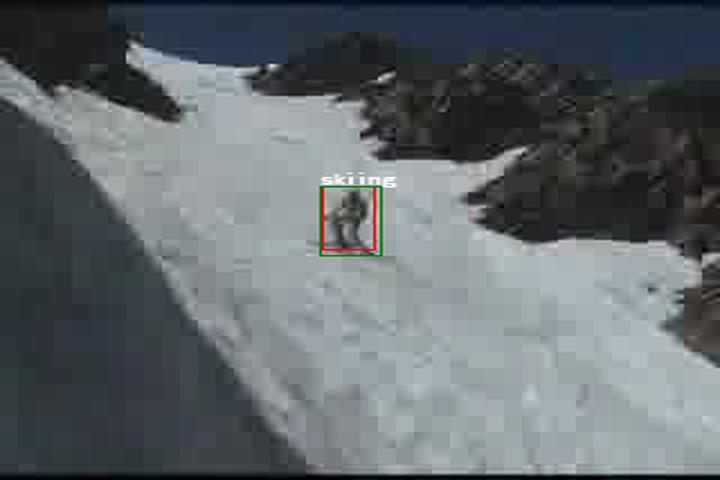} &
		\includegraphics[scale=0.15]{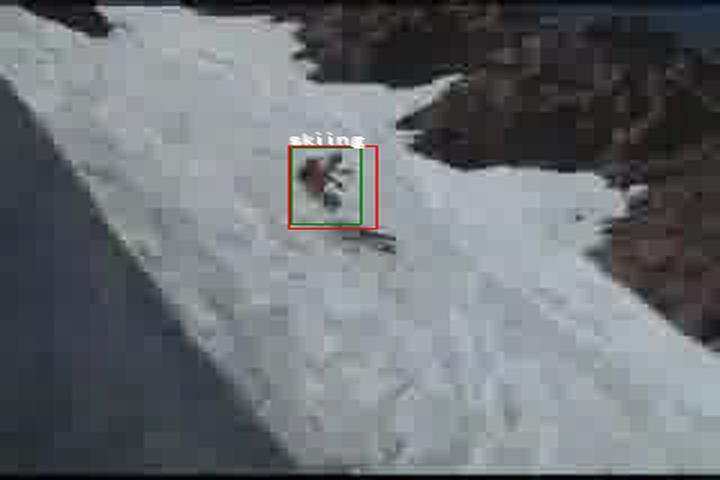} \\
		\rot{\scriptsize{Cliff Diving}} &  
		\includegraphics[scale=0.15]{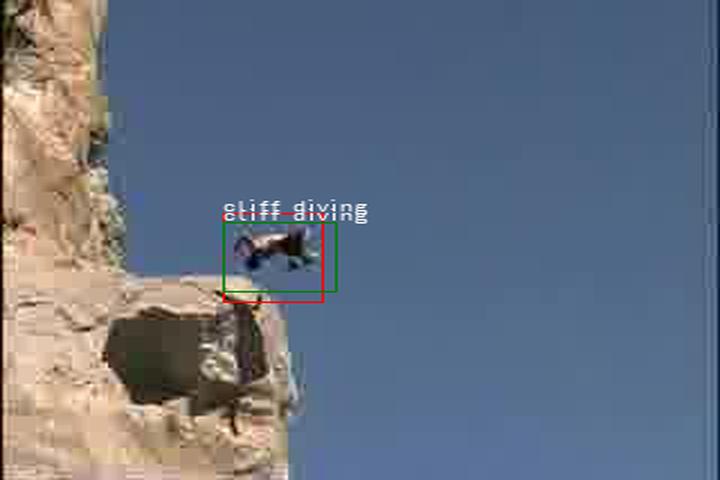}
		&
		\includegraphics[scale=0.15]{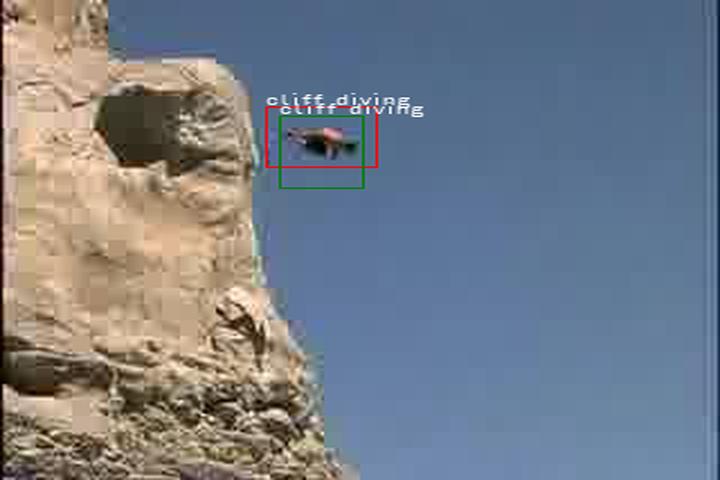} &
		\includegraphics[scale=0.15]{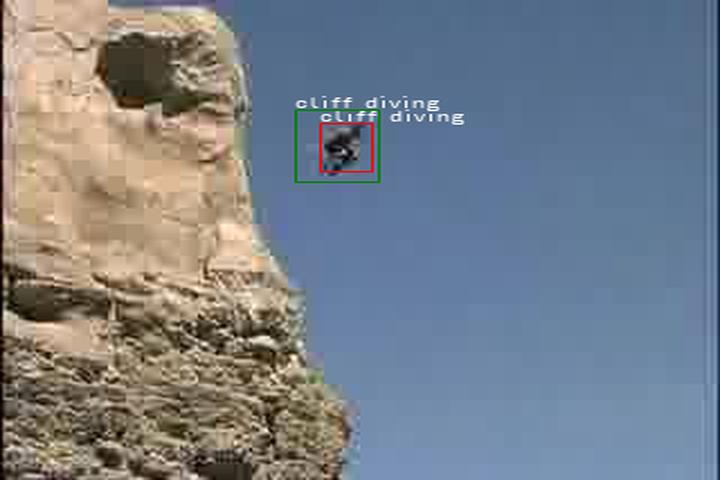} &
		\includegraphics[scale=0.15]{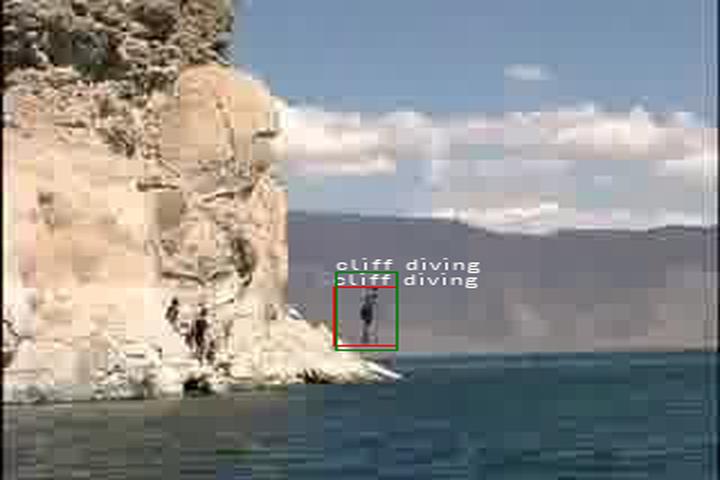} \\
	\end{tabularx}
	\label{samples}
	\caption{Action detection results for four action classes from the
		\textit{UCF-101}
		dataset using a model pre-trained using \textit{Kinetics}, and using
		tuned
		\textit{Flownet2} optical flow as input.}
\end{figure*}

\subsection{Comparison with Top Performers}

We compare our results with other top performers on the \textit{UCF-101-24}
dataset, as shown in Table.~\ref{top_performers}. It should be noted that out
of all reported results, only one variation of the Singh \textit{et al.}
framework runs in real-time (\SI{28}{fps}).

We observe that all of our models that use \textit{Kinetics} pre-training and
fine-tuning for \textit{Flownet2} variants outperform the other top performers.
However, we can only fairly compare our results to \citet{DBLP:journals/corr/HouCS17}, as both our tests use temporally trimmed
videos from the \textit{UCF-101} dataset. The other methods
\cite{DBLP:journals/corr/KalogeitonWFS17, DBLP:journals/corr/SinghSC16,
	DBLP:journals/corr/WeinzaepfelHS15, Peng2016} test on untrimmed videos, as
they perform both spatial and temporal detections. While they have an advantage
over our framework as linking actions temporally can improve the spatial
detections, they also suffer from a disadvantage as they have
a greater chance of getting a false positive if they detect an action in a frame
where there is no action being performed.

 \footnotetext[1]{As reported in \url{
 		https://github.com/gurkirt/realtime-action-detection }}

\begin{table}[!h]
\caption{Comparison of the f-mAP with other top performers using IoU
threshold of \(\alpha\).}
\begin{tabularx}{\columnwidth}{|X|P{0.4in}|}
\hline
 Model & \thead{\(\alpha\) = 0.5} \\
 \hline
 Weinzaepfel \textit{et al.} \cite{DBLP:journals/corr/WeinzaepfelHS15}
 \(\dagger\) & 35.84 \\
Hou \textit{et al.} \cite{DBLP:journals/corr/HouCS17} \(\star\) & 41.37 \\
Peng \textit{et al.} \cite{Peng2016}  \(\dagger\) & 65.37 \\
Singh \textit{et al.} \cite{DBLP:journals/corr/SinghSC16} RGB + DIS-Fast
\(\dagger\) \(\psi\) & 65.66\footnotemark[1] \footnotetext{As reported in
https://github.com/gurkirt/realtime-action-detection }\\
Singh \textit{et al.} \cite{DBLP:journals/corr/SinghSC16} RGB + Brox
\(\dagger\) & \textbf{68.31}\footnotemark[1]\\
Kalogeiton \textit{et al.}\cite{DBLP:journals/corr/KalogeitonWFS17}
\(\dagger\) & 67.1 \\

 \hline
Brox + Kinetics \(\star\) & 73.18 \\
Tuned Flownet2 + Kinetics \(\star\) & \textbf{74.07} \\
Tuned Flownet2-CSS + Kinetics \(\star\) \(\psi\) & 72.13\\
Tuned Flownet2-SD + Kinetics \(\star\)  \(\psi\) & 71.67 \\
\hline
\end{tabularx}
\raggedright \(\dagger\) : untrimmed videos.  \(\star\) : trimmed videos.
\(\psi\) : real-time .

\label{top_performers}
\end{table}

\subsection{Detection Runtime}

We propose an end-to-end trainable pipeline. Integrating the flow computation
in our framework using \textit{Flownet2} improves the compute resources
utilization.  We can make the best use of GPU parallelization in addition to
reducing the overhead caused by memory transfer if the framework is separated
into two parts. The frame per second (fps) rates for our architectures are shown in
Table.~\ref{runtime}. We used a NVIDIA GTX Titan X GPU for testing the runtime
speed which is the same card used for previously proposed work on real-time
action detection \cite{DBLP:journals/corr/SinghSC16}. We test using a batch
sizes of 1 and 4. With a batch size of 1 (online), the system
will have no latency. If a small latency is acceptable, we can buffer the input
frames to use a batch size of 4 which improves the frame per second
rate. We compare our results to \citet{DBLP:journals/corr/SinghSC16}, the
only real-time method for action detection. However, in their reported runtime,
they do not account for the overhead caused  by transferring the optical
flow computed using DIS-Fast to their two-stream SSD networks. Nevertheless,
our
model using
\textit{Flownet2-SD} is the fastest, achieving \SI{25}{fps} with no latency
or \SI{31}{fps} with minimal latency.

\begin{table}[!h]
	\caption{Frames per second rate of our models compared to the other reported
  real-time method.}
	\begin{tabularx}{\columnwidth}{|X|P{0.7in}|P{0.7in}|}
		\hline
		Model & batch size = 1 &  batch size = 4 \\
		\hline
		\citet{DBLP:journals/corr/SinghSC16} RGB+DIS-Fast & - & 28\\

		\hline

		Tuned Flownet2 + Kinetics  & 12 & 15 \\
		Tuned Flownet2-CSS + Kinetics & 17 & 21 \\
		Tuned Flownet2-SD + Kinetics &  \textbf{25} & \textbf{31} \\
		\hline
	\end{tabularx}
	\label{runtime}
\end{table}

\section{Conclusion}

In this work, we propose a real-time, end-to-end trainable two-stream network
for action detection by generalizing the \textit{YOLOv2} network architecture.
We train two-stream \textit{YOLOv2} networks jointly to learn
complementary features between the appearance and motion streams. We show that
transfer learning from the task of action recognition to action detection introduces
a boost in performance. Additionally, fine-tuning a trainable optical flow
estimator for the task of action detection results in a better representation
for the action-related motion in the scene, improving our model's performance.
Finally, we show that by integrating the optical flow computation and training
end-to-end, our framework runs in real-time (\SI{31}{fps}), faster than all
previous methods.

\section*{Acknowledgement}

We would like to thank Brendan Duke of the Machine Learning Research Group at
the University of Guelph for his help with training the \textit{Kinetics}
dataset and helpful suggestions toward improving the manuscript.

\bibliographystyle{unsrtnat}
\bibliography{references}

\end{document}